%% file: main.tex
\documentclass{Interspeech2024}

\interspeechcameraready 

\usepackage{graphicx}
\usepackage[table,xcdraw]{xcolor}
\usepackage{colortbl}
\usepackage{color}
\usepackage{multirow}
\usepackage[symbol*]{footmisc}

\title{Key-Element-Informed sLLM Tuning for Document Summarization}

\name[affiliation={1}]{Sangwon}{Ryu$^*$}
\name[affiliation={1}]{Heejin}{Do$^*$}
\name[affiliation={3}]{Yunsu}{Kim}
\name[affiliation={1,2}]{Gary Geunbae}{Lee}
\name[affiliation={1,2}]{Jungseul}{Ok$^\dagger$}

\address{
  $^1$Graduate School of Artificial Intelligence, POSTECH, South Korea\\
  $^2$Department of Computer Science and Engineering, POSTECH, South Korea \\
  $^3$aiXplain Inc., Los Gatos, CA, USA}
\email{\{ryusangwon, heejindo, gblee, jungseul\}@postech.ac.kr, yunsu.kim@aixplain.com}


\begin{document}

\maketitle

\let\oldthefootnote\thefootnote
\renewcommand{\thefootnote}{*}
\footnotetext{equal contribution}
\renewcommand{\thefootnote}{\arabic{footnote}}

\let\oldthefootnote\thefootnote
\renewcommand{\thefootnote}{$\dagger$}
\footnotetext{correspondence to: jungseul@postech.ac.kr}
\renewcommand{\thefootnote}{\arabic{footnote}}

\input{INTERSPEECH-2024/content/abstract}

\input{INTERSPEECH-2024/content/intro}

\input{INTERSPEECH-2024/content/related}

\input{INTERSPEECH-2024/content/method}

\input{INTERSPEECH-2024/content/experiment}

\input{INTERSPEECH-2024/content/conclusion}

\bibliographystyle{IEEEtran}
\bibliography{custom}

\end{document}

%% file: INTERSPEECH-2024/content/abstract.tex
\begin{abstract}
    Remarkable advances in large language models (LLMs) have enabled high-quality text summarization. However, this capability is currently accessible only through LLMs of substantial size or proprietary LLMs with usage fees. In response, smaller-scale LLMs (sLLMs) of easy accessibility and low costs have been extensively studied, yet they often suffer from missing key information and entities, i.e., low relevance, in particular, when input documents are long. We hence propose a key-element-informed instruction tuning for summarization, so-called \textit{KEITSum}, which identifies key elements in documents and instructs sLLM to generate summaries capturing these key elements. Experimental results on dialogue and news datasets demonstrate that sLLM with \textit{KEITSum} indeed provides high-quality summarization with higher relevance and less hallucinations, competitive to proprietary LLM.
\end{abstract}

%% file: INTERSPEECH-2024/content/intro.tex
\section{Introduction}

With the advent of Large Language Models (LLMs), recent studies have utilized LLMs across a broad spectrum of applications. Consequently, for summarization tasks, there is a paradigm shift from traditional encoder-decoder-based models \cite{vaswani2017attention, lewis2019bart, zhang2020pegasus, raffel2020t5, liu2022brio} to LLMs. It has been revealed that LLMs produce more contextual and natural summaries than the encoder-decoder models 
\cite{goyal2023news, zhang2023benchmarking, pu2023summarization} where LLMs do not merely put words from the document; instead, they substitute appropriate synonyms for a summary, resulting in more natural expressions and flows \cite{goyal2023news}. 
Noticeably, LLMs often generate even better summaries than human-written references \cite{zhang2023benchmarking, pu2023summarization}.

However, such a high-quality summarization has been only accessible by proprietary LLMs with usage fees or LLMs of large sizes. To improve accessibility, publicly available smaller-scale LLMs (sLLMs) can be considered. Noting that sLLMs can generate more fluent sentences than traditional encoder-decoder models, using sLLMs for summarization is a promising approach. However, according to our evaluation (Figure \ref{fig: output_ratio}), they still suffer from the problem of omitting key entities or information but including superfluous sentences in summaries, i.e., {\it low relevance}.



Hence, we aim to unleash the summarization capabilities of sLLMs by addressing the problem of low relevance. To this end, we propose key-element-informed sLLM tuning for document summarization (\textit{KEITSum}), of which an overview is illustrated in Figure~\ref{fig: main}. 
Given an input document, KEITSum identifies key elements consisting of the named entities and conclusion sentence and then instructs a fine-tuned sLLM to include the key elements when generating a summary, where the fine-tuning is conducted to optimize the sLLM for the key-element-informed summarization.
We evaluate KEITSum on a dialogue summarization dataset, DialogSum \cite{chen-etal-2021-dialogsum}, and a news summarization dataset, CNN/DM \cite{nallapati-etal-2016-abstractive}, using a multi-dimensional metric to assess summarization quality, \texttt{UniEval} \cite{zhong-etal-2022-towards}.
It is demonstrated that \textit{KEITSum} improves the summary quality compared to the baseline LLaMA2-7B, particularly in terms of \textit{relevance} and when summarizing long dialogs or documents.
In addition, we also observed that KEITSum is effective in reducing hallucinations.

%% file: INTERSPEECH-2024/content/related.tex
\input{INTERSPEECH-2024/fig/fig-main}

\section{Related work}

\textbf{Information omission in dialogue summarization.} Information or entity omission remains a persistent challenge in dialogue summarization. Traditional encoder-decoder models have tried to overcome this problem using various methods: \cite{zou2023understanding} introduces a method to detect information missing in conversations. \cite{tang-etal-2022-confit} introduces contrastive and self-supervised losses to address entity omission and other inconsistency problems. \cite{deutsch-roth-2023-incorporating} guided the inclusion of important spans identified through Question-Answering (QA) signals into the summaries. However, research on addressing missing information in dialogue datasets via sLLMs has not yet been extensively explored.




\noindent\textbf{Entity extraction for summarization.} Methods for extracting entities from the document to ensure their inclusion in the summaries have been introduced to mitigate entity omission in other summarization domains. \cite{berezin2023named} used the named entity recognition (NER) by masking extracted entities instead of random tokens when pre-training BART. However, it still has fundamental limitations inherent to encoder-decoder models. \cite{wang-etal-2023-element} employed a two-stage CoT method, where elements were extracted via GPT-3 \cite{NEURIPS2020_1457c0d6} in the initial stage, and then GPT-3 was utilized again to integrate those extracted elements to generate a summary. However, it could achieve element extraction only with models exceeding 175B parameters, requiring tremendous costs. Distinguished from their works, we aim to leverage the previously unexplored sLLM, LLaMA2-7B, to take advantage of its comprehending abilities while alleviating the cost burden. Unlike the API-relied entity extraction of \cite{wang-etal-2023-element}, our simple use of NER further diminishes the burden.

\noindent\textbf{Evaluation metrics.} Recently, critical limitations of the ROUGE score have been pointed out \cite{wang-etal-2023-element, scialom2021questeval, honovich-etal-2021-q2, zhang2023benchmarking}: it highly relies on the number of overlapping words and, thus, devalues appropriate synonyms generated in LLM. Furthermore, ROUGE is unable to evaluate entity omission or hallucination \cite{wan-etal-2023-faithfulness, roit-etal-2023-factually, goyal-durrett-2020-evaluating, kryscinski-etal-2020-evaluating, zhong-etal-2022-towards, bert-score, liu-etal-2023-g, ryu2024multidimensional}. Therefore, various multi-dimensional evaluation metrics have emerged \cite{scialom2021questeval, honovich-etal-2021-q2, zhong-etal-2022-towards, bert-score, liu-etal-2023-g}, among which \texttt{UniEval} is known to have the highest correlation with human evaluation currently. \texttt{UniEval} assesses scores for \textit{coherence}, \textit{consistency}, \textit{fluency}, and \textit{relevance}. We mainly aim to improve \textit{relevance}, which evaluates whether only the key information from the document has been included in the summary.

%% file: INTERSPEECH-2024/fig/fig-main.tex
\begin{figure*}[t]
\centering
\includegraphics[width=0.86\textwidth]{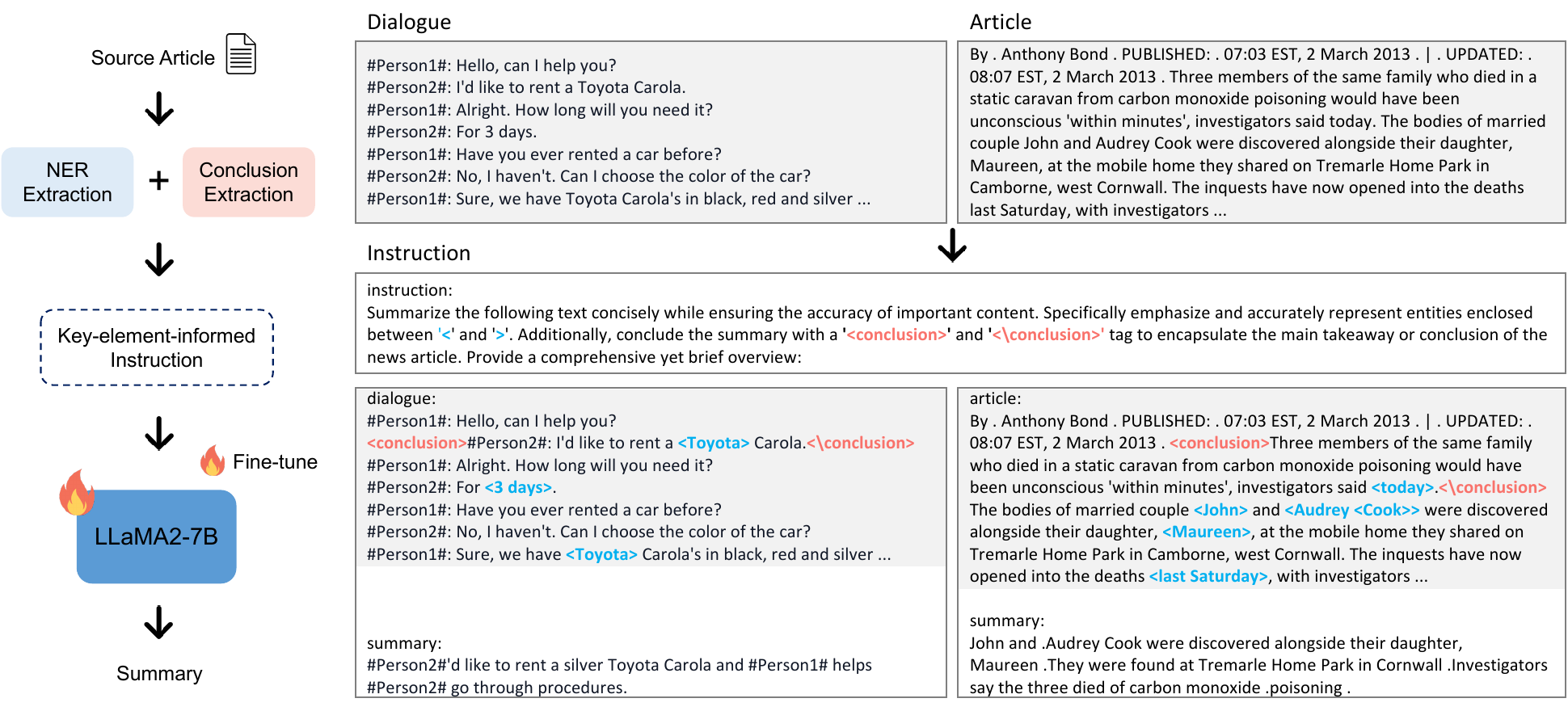}
\caption{Description of KEITSum Framework. We extract named entities and conclusion sentence from the source document and insert emphasis tokens. Following this, we create a full description by adding detailed instructions.}
\label{fig: main}
\end{figure*}

%% file: INTERSPEECH-2024/content/method.tex
\section{Key-element-informed tuning}
To efficiently capture the critical elements for the dialogue document, we propose a key-element-informed tuning for sLLMs. Specifically, we first extract two distinct key elements, identified as named entities and the conclusion sentence, using separate models. Then, we perform instruction tuning on the sLLM to guide the model in focusing on those extracted elements while generating the summary.

\subsection{Key-element extraction}

\noindent
\textbf{Entity extraction.} We use the NER mechanism for entity extraction. To select the named entities for extraction, we calculate the ratio of entities appearing in both the dialogues and the summaries. Table \ref{tab: entity_ratio} presents the proportion of named entities appearing in the dialogue that also appear in the reference summary. If a named entity appears in the reference summary with a high frequency, it indicates that such named entity should be included in the summary; therefore, we select named entities that appeared in more than 30\% of the summaries. Additionally, we conduct experiments with a news dataset and, following \cite{wang-etal-2023-element}, we use entities suitable for the news domain, such as \textit{person}, \textit{date}, \textit{organization}, and \textit{event}.

After extracting the entities suitable for each domain, we emphasize each entity in the document by surrounding them with the emphasis tokens, \texttt{<} and \texttt{>}. Unlike a prior work \cite{wang-etal-2023-element}, our approach solely emphasizes the entities with tokens without explicitly listing their meaning. 

\input{INTERSPEECH-2024/fig/tab-entity_ratio}

\noindent
\textbf{Conclusion extraction.} Furthermore, to extract the key sentence from the document, we employ a pre-trained BERT-based extractive summarizer \cite{miller2019leveraging} and select the top-1 sentence. This is motivated by combining extractive summarization with abstractive methods to improve summary quality \cite{mao2021constrained, liu2019text}. Instead of explicitly passing the selected sentence, we merely mark the sentence in the document when designing the instruction. Specifically, we highlight the key sentence by adding a distinct token by encapsulating it between \texttt{<conclusion>} and \texttt{</conclusion>} tokens. A more concentrated summary can be generated by implicitly guiding the model to conclude the summary using the marked main points. Figure \ref{fig: main} illustrates the overall structure.

\input{INTERSPEECH-2024/fig/tab-main-dialog}

\input{INTERSPEECH-2024/fig/tab-length}

\subsection{Instruction tuning} 
As a prompt for fine-tuning the sLLM, we provide the instruction with a key-element-informed document and a reference summary (Figure \ref{fig: main}). For the instruction, we describe a task definition and explain how the key elements are emphasized in the following source document. Addressing missing information or entities can potentially lead to hallucinations \cite{zou2023understanding}; thus, we sought to mitigate this trade-off by explicitly demanding \textit{accurate} generation in detailed instructions. In particular, we concatenate the instruction ($i$), converted document ($d'$), and reference summary ($s$) to construct the prompt ($[i; d';s]$). Then, we fine-tune the sLLM using the designed prompt. This key-element-informed tuning enables the model to focus more on important points within the document in the generation process.



%% file: INTERSPEECH-2024/fig/tab-entity_ratio.tex
\definecolor{lightblue}{RGB}{239,247,255} 

\begin{table}
\caption{The numbers in Dialogue and Summary represent the count of samples containing each entity out of 500 in the validation set. The ratio is the number in the Summary divided by that in the Dialogue. Blue background highlights selected entities.}
\centering
\scalebox{0.65}
{\begin{tabular}{l|ccc}
\hline
 {\small Named Entity} & {\small Ratio} & {\small Dialogue} & {\small Summary} \\ 
 \hline

\rowcolor{lightblue}
PERSON  & 0.839 & 186 & 156  \\ 
\rowcolor{lightblue}
GPE  & 0.481 & 81 & 39  \\ 
\rowcolor{lightblue}
LANGUAGE  & 0.474 & 19 & 9  \\
\rowcolor{lightblue}
ORG  & 0.411 & 56 & 23  \\ 
\rowcolor{lightblue}
FAC   & 0.350 & 20 & 7  \\ 
\rowcolor{lightblue}
NORP  & 0.333 & 42 & 14 \\
\rowcolor{lightblue}
DATE  & 0.311 & 183 & 57  \\
MONEY  & 0.182 & 55 & 10  \\
ORDINAL  & 0.180 & 50 & 9  \\
CARDINAL  & 0.172 & 145 & 25  \\
TIME  & 0.143 & 112 & 16  \\
LOC  & 0.071 & 14 & 1  \\

\hline
\end{tabular}}
\label{tab: entity_ratio}
\end{table}




%% file: INTERSPEECH-2024/fig/tab-main-dialog.tex


\begin{table*}
\caption{Comparison between encoder-decoder-based models, LLaMA-2-7B, and GPT-3 in DialogSum and CNN/DM dataset. KEITSum$_{all}$ refers to the results when all entities are extracted, while KEITSum$_{top-1}$ indicates the results when only the entity with the highest proportion is extracted. Finally, KEITSum represents the outcomes when entities with a ratio of over 30\% are extracted.}
\centering
\scalebox{0.8}{
\begin{tabular}{l|l|ccccc|c}
\hline
\multirow{2}{*}{Dataset} & \multirow{2}{*}{Model} & \multicolumn{5}{c|}{UniEval} & \multirow{2}{*}{ROUGE-1} \\ \cline{3-7} &  & Coherence & Consistency & Fluency & \multicolumn{1}{c|}{Relevance} & Overall &  \\ \hline
\multirow{7}{*}{DialogSum} & BART \cite{lewis2019bart}  & 0.928 & 0.914 & 0.913 & \multicolumn{1}{c|}{0.846} & 0.900 & 0.418 \\
 & T-5 \cite{raffel2020t5} & 0.948 & 0.939 & 0.920 & \multicolumn{1}{c|}{0.870} & 0.919 & 0.414 \\ \cline{2-8} 
 &  LLaMA2-7B (Fine-tuned)  & 0.959 & 0.939 & 0.935 & \multicolumn{1}{c|}{0.912} & 0.936 & \textbf{0.440} \\ 
 & \textbf{KEITSum$_{all}$} (ours)  & 0.963 & \textbf{0.942} & 0.939 & \multicolumn{1}{c|}{0.914} & 0.939 & 0.429 \\
 & \textbf{KEITSum$_{top-1}$} (ours) & 0.962 & 0.941 & 0.941 & \multicolumn{1}{c|}{0.915} & 0.940 & 0.429 \\
 & \textbf{KEITSum} (ours)  & \textbf{0.965} & \textbf{0.942} & \textbf{0.942} & \multicolumn{1}{c|}{\textbf{0.918}} & \textbf{0.942} & 0.430\\ \cline{2-8} 
 & \textcolor{gray}{GPT-3} \cite{NEURIPS2020_1457c0d6} & \textcolor{gray}{0.969} & \textcolor{gray}{0.907} & \textcolor{gray}{0.944} & \multicolumn{1}{c|}{\textcolor{gray}{0.932}} & \textcolor{gray}{0.938} & \textcolor{gray}{0.270}  \\ \hline
\multirow{8}{*}{CNN/DM} & PEGASUS \cite{zhang2020pegasus} & 0.944 & 0.935 & 0.829 & \multicolumn{1}{c|}{0.697} & 0.851 & 0.412 \\
 & BART \cite{lewis2019bart} & 0.956 & 0.943 & 0.837 & \multicolumn{1}{c|}{0.684} & 0.855 & 0.415 \\ 
 & T5 \cite{raffel2020t5} & \textbf{0.968} & \textbf{0.959} & 0.838 & \multicolumn{1}{c|}{0.767} & 0.883 & 0.426 \\
 & BRIO \cite{liu2022brio}  & 0.947 & 0.927 & 0.833 & \multicolumn{1}{c|}{0.790} & 0.874 & \textbf{0.455} \\ \cline{2-8} 
 & LLaMA2-7B (Fine-tuned) & 0.943 & 0.933 & 0.870 & \multicolumn{1}{c|}{0.770} & 0.879 & 0.364 \\ 
 & \textbf{KEITSum} (ours)  & 0.954 & 0.923 & \textbf{0.930} & \multicolumn{1}{c|}{\textbf{0.805}} & \textbf{0.903} & 0.343\\ \cline{2-8} 
 & \textcolor{gray}{GPT-3} \cite{NEURIPS2020_1457c0d6} & \textcolor{gray}{0.964} & \textcolor{gray}{0.909} & \textcolor{gray}{0.949} & \multicolumn{1}{c|}{\textcolor{gray}{0.905}} & \textcolor{gray}{0.932} & \textcolor{gray}{0.399} \\ 
 & \textcolor{gray}{GPT-3 + CoT} \cite{wang-etal-2023-element} & \textcolor{gray}{0.948} & \textcolor{gray}{0.870} & \textcolor{gray}{0.948} & \multicolumn{1}{c|}{\textcolor{gray}{0.910}} & \textcolor{gray}{0.919} & \textcolor{gray}{0.464} \\ \hline
\end{tabular}
}
\label{tab: UniEval_dialogsum}
\end{table*}

%% file: INTERSPEECH-2024/fig/tab-length.tex
\begin{table*}
\caption{The performance variation of KEITSum on the DialogSum according to dialogue length. The test set was divided based on the average length of dialogues.}
\centering
\scalebox{0.78}
{\begin{tabular}{l|c|cc|ccccc}
\hline
\multirow{2}{*}{Model} & \multirow{2}{*}{\# of Dialogues} & \multicolumn{2}{c|}{Summary Length} & \multicolumn{5}{c}{UniEval} \\ \cline{3-9}
 & & \multicolumn{1}{c|}{Document} & Summary & Coherence & Consistency & Fluency & \multicolumn{1}{c|}{Relevance} & Overall \\ \hline
LLaMA2-7B$_{short}$ & \multirow{2}{*}{882} & \multicolumn{1}{c|}{\multirow{2}{*}{87.6}} & 17.2 & 0.961 & 0.948 & 0.933 & \multicolumn{1}{c|}{0.914} & 0.939 \\
KEITSum$_{short}$ & & \multicolumn{1}{c|}{} & 20.0 & 0.964 & 0.948 & 0.939 & \multicolumn{1}{c|}{0.916} & 0.942 \\ \hline
LLaMA2-7B$_{long}$ & \multirow{2}{*}{618} & \multicolumn{1}{c|}{\multirow{2}{*}{201.3}} & 32.5 & 0.955 & 0.927 & 0.938 & \multicolumn{1}{c|}{0.909} & 0.932 \\
KEITSum$_{long}$ & & \multicolumn{1}{c|}{} & 35.6 & 0.965 & 0.934 & 0.945 & \multicolumn{1}{c|}{0.920} & 0.941 \\ \hline
\end{tabular}}
\label{tab: length_ablation}
\end{table*}

%% file: INTERSPEECH-2024/content/experiment.tex
\section{Experimental setup}

\noindent
\textbf{Datasets.}
We use the DialogSum dataset \cite{chen-etal-2021-dialogsum}, a large-scale dialogue summarization dataset. It comprises a 12.5K training set and a 1.5K test set, each accompanied by a human-written summary that captures the most salient information and entities. It encompasses a broad spectrum of daily-life topics through face-to-face spoken dialogues with a diverse distribution of lengths. To demonstrate domain extensibility, we employ the CNN/Daily Mail (CNN/DM) dataset \cite{nallapati-etal-2016-abstractive}, a news article collection paired with multi-sentence human-written summaries. In contrast to the encoder-decoder model trained on the full dataset, the sLLMs were trained on only 10,000 subsets for efficiency in both datasets. Following previous research that highlights the poor quality of reference summaries in the CNN/DM \cite{zhang2023benchmarking}, we use the recently released \textit{element-aware} test set \cite{wang-etal-2023-element} designed to address the deficiencies of the original dataset.

\noindent\textbf{Models.} To extract entity, we use the Flair\footnote{https://github.com/flairNLP/flair} \cite{akbik2019flair}, a well-designed NER framework. It was specifically pre-trained on the \textit{OntoNote5} \cite{pradhan2013towards} for NER tasks in various domains, such as conversational speech and broadcast. For key sentence extraction, we use the BERT summarizer\footnote{https://pypi.org/project/bert-extractive-summarizer/} \cite{miller2019leveraging}. For the sLLM, we fine-tune the smallest LLaMA2 \cite{touvron2023llama} of 7 billion parameters, one of the famous open-source sLLMs. We fine-tune both LLaMA2 and KEITSum via LoRA \cite{hu2021lora}, which facilitates efficient training by modifying a limited parameter subset while original ones are frozen; thus, it eliminates the need for full-model retraining. We fine-tune LLaMA2 using a basic prompt format commonly used for summarization tasks.
We set rank $r$=$8$, $dropout$=$0.05$, $alpha$=$32$, and $epoch$=$3$ as LoRA hyperparameter.
As our comparative models, we use robust encoder-decoder models, such as BART \cite{lewis2019bart}, T5 \cite{raffel2020t5}, PEGASUS \cite{zhang2020pegasus} and BRIO \cite{liu2022brio}. They were fine-tuned on the entire training set. For GPT-3 \cite{NEURIPS2020_1457c0d6}, we generated summaries using \texttt{GPT-3.5-turbo} for DialogSum, while we used summaries created by \cite{wang-etal-2023-element} using the \texttt{text-davinci-002} for CNN/DM.

\noindent
\textbf{Evaluation metrics.} ROUGE scores, which fail to evaluate summaries properly \cite{wan-etal-2023-faithfulness, roit-etal-2023-factually, goyal-durrett-2020-evaluating, kryscinski-etal-2020-evaluating, zhong-etal-2022-towards, bert-score, liu-etal-2023-g}, suffer from another significant drawback: heavy reliance on reference summaries. Recent research highlighted that the quality of reference summaries in abstractive summarization is often subpar \cite{zhang2023benchmarking, adams-etal-2022-learning}.

Thus, to measure the omission and hallucination of the summaries precisely, we employ \texttt{UniEval} \cite{zhong-etal-2022-towards} and human evaluation for multi-dimensional evaluation, and ChatGPT evaluation \cite{chiang2023can} to examine the presence of inconsistencies in the summaries. \texttt{UniEval} is a recently proposed multi-dimensional evaluation tool for natural language generation (NLG) tasks, which demonstrates the highest correlation with human evaluation among open-source multi-dimensional evaluation metrics. While not overly relying on reference summaries, it provides four explainable evaluation dimensions: \textit{coherence}, \textit{consistency}, \textit{fluency}, and \textit{relevance}. To gauge the extent of hallucinations in model-generated summaries, we use the recently introduced ChatGPT Evaluation \cite{chiang2023can, shen2023large}. Finally, we conduct the human evaluation.

\input{INTERSPEECH-2024/fig/fig-ratio}

\section{Results and discussions}

\subsection{Main results}

\textbf{Multi-dimensional evaluation.} As shown in Table \ref{tab: UniEval_dialogsum}, our approach demonstrated improvements across all \texttt{UniEval} dimensions in the DialogSum. Emphasizing only the high-relevance named entities, rather than highlighting all named entities as done in KEITSum$_{all}$ or the most frequent named entity as done in KEITSum$_{top-1}$, slightly benefited performance enhancement. Notably, by ensuring the inclusion of essential elements in the summaries, KEITSum boosted \textit{relevance} score in both the DialogSum and CNN/DM. As a result, our model achieved higher overall scores not only compared to existing encoder-decoder-based summarization models but also comparable to the much larger model, GPT-3.

Compared to the encoder-decoder models fine-tuned with the full dataset in the CNN/DM dataset, our model performs better in most dimensions despite using only 3.6\% of the train set. In detail, it shows lower and \textit{consistency} scores while exhibiting markedly higher \textit{fluency} scores. This could be attributed to the difference in the generation procedure, i.e., encoder-decoder models often generate content directly from the source text, resulting in high \textit{consistency}, whereas our decoder-only approach leads to diverse yet more appropriate synonyms in the summaries. Even GPT-3 and GPT-3+CoT show comparable or lower scores on these aspects than T5, thereby supporting our assumption.

Additionally, we measured the ROUGE-1 scores for each model. Table \ref{tab: UniEval_dialogsum} showed that the ROUGE-1 score of KEITSum slightly decreased compared to the LLaMA2-7B model on DialogSum. Moreover, the GPT-3 model, known for generating the highest quality summaries, showed lower ROUGE-1 scores than other summarization models. This underscores once again that ROUGE scores are insufficient to measure the quality of summaries generated by LLMs and fail to capture dimensions such as \textit{relevance}.

\input{INTERSPEECH-2024/fig/tab-human}

\noindent
\textbf{Entity ratio.} To verify the actual inclusion of the emphasized entities in the generated summaries, we investigate the entity ratio using the CNN/DM element-aware test set \cite{wang-etal-2023-element}. We extracted the entities in the reference summaries and then calculated the ratio of these entities present in the summaries produced by each model. Figure~\ref{fig: output_ratio} shows that \textit{KEITSum} measured similarly to the tendencies of GPT-3 or GPT-3+CoT, exhibiting a notable improvement over LLaMA2-7B across all entities. Remarkably, the ratio of EVENT entities shows a considerable increase, where the LLaMA2-7B notably failed to capture well.

\subsection{Length dependency}
When the document is longer, more frequent missing information issues occur. Therefore, our method, emphasizing entities and key sentences to ensure accurate entities are included in the summary, is more effective in longer text. Indeed, as seen in Tables \ref{tab: UniEval_dialogsum}, there is a greater performance improvement in the CNN/DM, which has a longer average text length than DialogSum. For a more detailed analysis, we divide the DialogSum dataset into long and short categories based on the average length of the text. As shown in Table \ref{tab: length_ablation}, while there was a slight performance improvement when the summary length was short, our model showed notable performance improvement when the summary length was long.


\subsection{Human evaluation}
We conducted a human evaluation to ascertain the performance improvement of our model compared to other summarization models (Table \ref{tab: human}). We hired three English teachers to assess 20 dialogues via Upwork\footnote{https://www.upwork.com/}. The evaluation criteria encompass \textit{comprehension}, \textit{faithfulness}, \textit{relevance}, \textit{fluency}, and \textit{overall} score based on individual preference, rated on a scale of 0 to 5 (highest). KEITSum surpassed LLaMA2-7B in \textit{faithfulness} and \textit{relevance}, reflecting better alignment with the original document and inclusion of only crucial information. This improvement stems from our focus on key entities and sentences, ensuring no important details are missed in the summaries.




\input{INTERSPEECH-2024/fig/fig-hallucination}

\subsection{Measuring hallucinations}


As Incorporating missing entities can potentially lead to hallucinations \cite{zou2023understanding}, we quantified how inconsistent information was present in the generated summaries. Inspired by the following research findings that ChatGPT can evaluate in a manner similar to humans \cite{gao2023humanlike, chiang2023can, du2023quantifying}, we employed ChatGPT to gauge the extent of hallucination in model-generated summaries of 20 dialogue samples; here, hallucination refers to any incorrect content, including misattribution, misinterpretation, and redundant content. Figure \ref{fig: hallucination} illustrates that our model produces summaries with an average of 60\% fewer hallucinations per dialogue than those generated by LLaMA2-7B, even surpassing the reference summaries in hallucination reduction.




%% file: INTERSPEECH-2024/fig/fig-ratio.tex
\begin{figure}
\centering
\includegraphics[width=0.33 \textwidth]{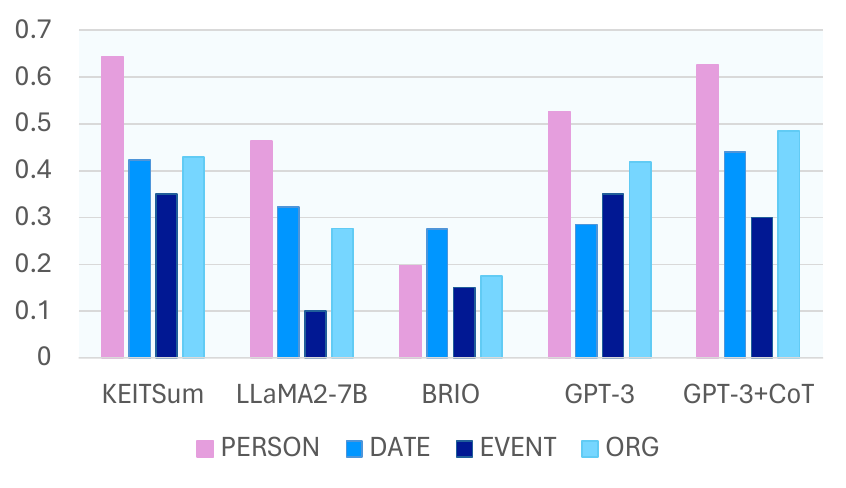}
\caption{The proportion of entities included in the element-aware dataset that are also included in the summaries generated by each model.}
\label{fig: output_ratio}
\end{figure}

%% file: INTERSPEECH-2024/fig/tab-human.tex

\begin{table}
\caption{Human evaluation in DialogSum.}
\centering
\scalebox{0.68}{
\begin{tabular}{l|cccc|c}
\hline
Model & \multicolumn{1}{l}{Comprehension} & \multicolumn{1}{l}{Faithfulness} & \multicolumn{1}{l}{Relevance} & \multicolumn{1}{l|}{Fluency} & \multicolumn{1}{l}{Overall} \\ \hline
BART & 4.192 & 3.440 & 3.652 & 4.368 & 3.910 \\
T5 & 4.130 & 3.322 & 3.527 & 4.288 & 3.838 \\
Reference & 4.363 & 3.812 & 3.950 & 4.460 & 4.137 \\
LLaMA2-7B & 4.413 & 3.927 & 4.040 & 4.458 & 4.183 \\
KEITSum & \textbf{4.527} & \textbf{4.115} & \textbf{4.145} & \textbf{4.562} & \textbf{4.347} \\ \hline
\textcolor{gray}{GPT-3} & \textcolor{gray}{4.880} & \textcolor{gray}{4.740} & \textcolor{gray}{4.772} & \textcolor{gray}{4.876} & \textcolor{gray}{4.802} \\ \hline
\end{tabular}
}\label{tab: human}
\end{table}

%% file: INTERSPEECH-2024/fig/fig-hallucination.tex
\begin{figure}
\centering
\includegraphics[width=0.29\textwidth]{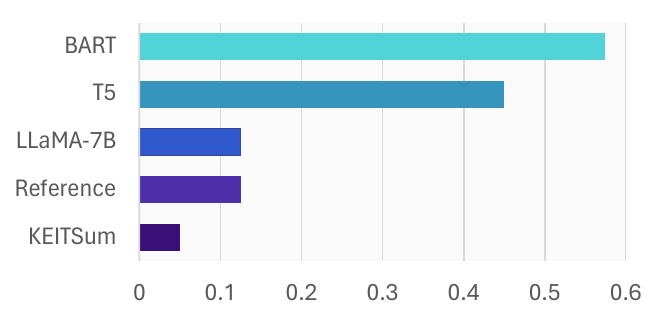}
\caption{Hallucination ratio per dialog in DialogSum.}
\label{fig: hallucination}
\end{figure}

%% file: INTERSPEECH-2024/content/conclusion.tex
\section{Conclusion}

With the advent of GPT-3, LLM-utilized summarization has achieved superior performance. However, large-scale proprietary LLMs are only accessible via APIs and are expensive, while the smaller public model, sLLMs, still struggles with entity omission in summarization and delivers inferior performance. We propose a key-element-informed instruction tuning method to overcome this issue in sLLMs. By adding emphasis tokens to essential elements and detailed instruction for summarization, \texttt{UniEval} scores noticeably improved in \textit{relevance}, exhibiting a comparable overall score of GPT-3. Furthermore, both 60\% reduced hallucinations on ChatGPT evaluation and 4.8\% improved faithfulness in human evaluation, proving the efficacy of our method.

\section{Acknowledgements} This work was supported by the National Research Foundation of Korea (NRF) grant funded by the Korea government (MSIT) (No. RS-2023-00217286) and Institute of Information \& communications Technology Planning \& Evaluation (IITP) grant funded by the Korea government (MSIT) (No.RS-2019-II191906, Artificial Intelligence Graduate School Program (POSTECH)).

%% file: main.bbl
\begin{thebibliography}{10}
\providecommand{\url}[1]{#1}
\csname url@samestyle\endcsname
\providecommand{\newblock}{\relax}
\providecommand{\bibinfo}[2]{#2}
\providecommand{\BIBentrySTDinterwordspacing}{\spaceskip=0pt\relax}
\providecommand{\BIBentryALTinterwordstretchfactor}{4}
\providecommand{\BIBentryALTinterwordspacing}{\spaceskip=\fontdimen2\font plus
\BIBentryALTinterwordstretchfactor\fontdimen3\font minus \fontdimen4\font\relax}
\providecommand{\BIBforeignlanguage}[2]{{%
\expandafter\ifx\csname l@#1\endcsname\relax
\typeout{** WARNING: IEEEtran.bst: No hyphenation pattern has been}%
\typeout{** loaded for the language `#1'. Using the pattern for}%
\typeout{** the default language instead.}%
\else
\language=\csname l@#1\endcsname
\fi
#2}}
\providecommand{\BIBdecl}{\relax}
\BIBdecl

\bibitem{vaswani2017attention}
Vaswani \emph{et~al.}, ``Attention is all you need,'' \emph{Advances in neural information processing systems}, 2017.

\bibitem{lewis2019bart}
M.~Lewis \emph{et~al.}, ``{BART}: Denoising sequence-to-sequence pre-training for natural language generation, translation, and comprehension,'' in \emph{Proceedings of the 58th Annual Meeting of the Association for Computational Linguistics}, 2020.

\bibitem{zhang2020pegasus}
Zhang \emph{et~al.}, ``Pegasus: pre-training with extracted gap-sentences for abstractive summarization,'' in \emph{Proceedings of the 37th International Conference on Machine Learning}, 2020.

\bibitem{raffel2020t5}
C.~Raffel \emph{et~al.}, ``Exploring the limits of transfer learning with a unified text-to-text transformer,'' \emph{Journal of Machine Learning Research}, 2020.

\bibitem{liu2022brio}
Y.~Liu \emph{et~al.}, ``{BRIO}: Bringing order to abstractive summarization,'' in \emph{Proceedings of the 60th Annual Meeting of the Association for Computational Linguistics}, 2022.

\bibitem{goyal2023news}
T.~Goyal \emph{et~al.}, ``News summarization and evaluation in the era of gpt-3,'' 2023.

\bibitem{zhang2023benchmarking}
T.~Zhang \emph{et~al.}, ``Benchmarking large language models for news summarization,'' 2023.

\bibitem{pu2023summarization}
X.~Pu, M.~Gao, and X.~Wan, ``Summarization is (almost) dead,'' \emph{arXiv preprint arXiv:2309.09558}, 2023.

\bibitem{chen-etal-2021-dialogsum}
Y.~Chen \emph{et~al.}, ``{D}ialog{S}um: {A} real-life scenario dialogue summarization dataset,'' in \emph{Findings of the Association for Computational Linguistics: ACL-IJCNLP 2021}, 2021.

\bibitem{nallapati-etal-2016-abstractive}
R.~Nallapati \emph{et~al.}, ``Abstractive text summarization using sequence-to-sequence {RNN}s and beyond,'' in \emph{Proceedings of the 20th {SIGNLL} Conference on Computational Natural Language Learning}, 2016.

\bibitem{zhong-etal-2022-towards}
M.~Zhong \emph{et~al.}, ``Towards a unified multi-dimensional evaluator for text generation,'' in \emph{Proceedings of the 2022 Conference on Empirical Methods in Natural Language Processing}, 2022.

\bibitem{zou2023understanding}
Y.~Zou \emph{et~al.}, ``Towards understanding omission in dialogue summarization,'' in \emph{Proceedings of the 61st Annual Meeting of the Association for Computational Linguistics (Volume 1: Long Papers)}, 2023.

\bibitem{tang-etal-2022-confit}
X.~Tang \emph{et~al.}, ``{CONFIT}: Toward faithful dialogue summarization with linguistically-informed contrastive fine-tuning,'' in \emph{Proceedings of the 2022 Conference of the North American Chapter of the Association for Computational Linguistics: Human Language Technologies}, Seattle, United States, 2022.

\bibitem{deutsch-roth-2023-incorporating}
D.~Deutsch and D.~Roth, ``Incorporating question answering-based signals into abstractive summarization via salient span selection,'' in \emph{Proceedings of the 17th Conference of the European Chapter of the Association for Computational Linguistics}, 2023.

\bibitem{berezin2023named}
S.~Berezin \emph{et~al.}, ``Named entity inclusion in abstractive text summarization,'' in \emph{Proceedings of the Third Workshop on Scholarly Document Processing}, 2022.

\bibitem{wang-etal-2023-element}
Y.~Wang \emph{et~al.}, ``Element-aware summarization with large language models: Expert-aligned evaluation and chain-of-thought method,'' in \emph{Proceedings of the 61st Annual Meeting of the Association for Computational Linguistics}, 2023.

\bibitem{NEURIPS2020_1457c0d6}
T.~Brown \emph{et~al.}, ``Language models are few-shot learners,'' in \emph{Advances in Neural Information Processing Systems}.\hskip 1em plus 0.5em minus 0.4em\relax Curran Associates, Inc., 2020.

\bibitem{scialom2021questeval}
T.~Scialom \emph{et~al.}, ``{Q}uest{E}val: Summarization asks for fact-based evaluation,'' in \emph{Proceedings of the 2021 Conference on Empirical Methods in Natural Language Processing}, 2021.

\bibitem{honovich-etal-2021-q2}
O.~Honovich \emph{et~al.}, ``$q^{2}$: {E}valuating factual consistency in knowledge-grounded dialogues via question generation and question answering,'' in \emph{Proceedings of the 2021 Conference on Empirical Methods in Natural Language Processing}, 2021.

\bibitem{wan-etal-2023-faithfulness}
D.~Wan \emph{et~al.}, ``Faithfulness-aware decoding strategies for abstractive summarization,'' in \emph{Proceedings of the 17th Conference of the European Chapter of the Association for Computational Linguistics}, May 2023.

\bibitem{roit-etal-2023-factually}
P.~Roit \emph{et~al.}, ``Factually consistent summarization via reinforcement learning with textual entailment feedback,'' in \emph{Proceedings of the 61st Annual Meeting of the Association for Computational Linguistics (Volume 1: Long Papers)}, 2023.

\bibitem{goyal-durrett-2020-evaluating}
T.~Goyal \emph{et~al.}, ``Evaluating factuality in generation with dependency-level entailment,'' in \emph{Findings of the Association for Computational Linguistics: EMNLP 2020}.\hskip 1em plus 0.5em minus 0.4em\relax Association for Computational Linguistics, 2020.

\bibitem{kryscinski-etal-2020-evaluating}
W.~Kryscinski \emph{et~al.}, ``Evaluating the factual consistency of abstractive text summarization,'' in \emph{Proceedings of the 2020 Conference on Empirical Methods in Natural Language Processing (EMNLP)}, 2020.

\bibitem{bert-score}
T.~Zhang* \emph{et~al.}, ``Bertscore: Evaluating text generation with bert,'' in \emph{International Conference on Learning Representations}, 2020.

\bibitem{liu-etal-2023-g}
Y.~Liu \emph{et~al.}, ``{G}-eval: {NLG} evaluation using gpt-4 with better human alignment,'' in \emph{Proceedings of the 2023 Conference on Empirical Methods in Natural Language Processing}, 2023.

\bibitem{ryu2024multidimensional}
S.~Ryu \emph{et~al.}, ``Multi-dimensional optimization for text summarization via reinforcement learning,'' in \emph{Proceedings of the 62nd Annual Meeting of the Association for Computational Linguistics}, 2024.

\bibitem{miller2019leveraging}
D.~Miller, ``Leveraging bert for extractive text summarization on lectures,'' 2019.

\bibitem{mao2021constrained}
Y.~Mao \emph{et~al.}, ``Constrained abstractive summarization: Preserving factual consistency with constrained generation,'' 2021.

\bibitem{liu2019text}
Y.~Liu \emph{et~al.}, ``Text summarization with pretrained encoders,'' in \emph{Proceedings of the 2019 Conference on Empirical Methods in Natural Language Processing and the 9th International Joint Conference on Natural Language Processing}, 2019.

\bibitem{akbik2019flair}
A.~Akbik, Bergmann \emph{et~al.}, ``{FLAIR}: An easy-to-use framework for state-of-the-art {NLP},'' in \emph{Annual Conference of the North American Chapter of the Association for Computational Linguistics (Demonstrations)}, 2019.

\bibitem{pradhan2013towards}
S.~Pradhan \emph{et~al.}, ``Towards robust linguistic analysis using ontonotes,'' in \emph{Proceedings of the Seventeenth Conference on Computational Natural Language Learning}, 2013.

\bibitem{touvron2023llama}
Touvron \emph{et~al.}, ``Llama 2: Open foundation and fine-tuned chat models,'' \emph{arXiv preprint arXiv:2307.09288}, 2023.

\bibitem{hu2021lora}
E.~Hu \emph{et~al.}, ``Lora: Low-rank adaptation of large language models,'' \emph{arXiv preprint arXiv:2106.09685}, 2021.

\bibitem{adams-etal-2022-learning}
\BIBentryALTinterwordspacing
G.~Adams \emph{et~al.}, ``Learning to revise references for faithful summarization,'' in \emph{Findings of the Association for Computational Linguistics: EMNLP 2022}, 2022. [Online]. Available: \url{https://aclanthology.org/2022.findings-emnlp.296}
\BIBentrySTDinterwordspacing

\bibitem{chiang2023can}
C.-H. Chiang and oth, ``Can large language models be an alternative to human evaluations?'' in \emph{Proceedings of the 61st Annual Meeting of the Association for Computational Linguistics (Volume 1: Long Papers)}, 2023.

\bibitem{shen2023large}
C.~Shen \emph{et~al.}, ``Are large language models good evaluators for abstractive summarization?'' \emph{arXiv preprint arXiv:2305.13091}, 2023.

\bibitem{gao2023humanlike}
M.~Gao, J.~Ruan, R.~Sun, X.~Yin, S.~Yang, and X.~Wan, ``Human-like summarization evaluation with chatgpt,'' \emph{arXiv preprint arXiv:2304.02554}, 2023.

\bibitem{du2023quantifying}
L.~Du \emph{et~al.}, ``Quantifying and attributing the hallucination of large language models via association analysis,'' 2023.

\end{thebibliography}
